\documentclass[letterpaper, 10 pt, conference]{packages/ieeeconf}
\IEEEoverridecommandlockouts

\usepackage[margin=0.75in]{geometry}
\usepackage{packages/algorithm}
\usepackage{packages/algorithmic}

\usepackage[compress]{cite}
\makeatletter
\let\NAT@parse\undefined
\makeatother
\usepackage[colorlinks=true]{hyperref}

\usepackage[table]{xcolor} \usepackage[section]{placeins} 

\usepackage{graphics} \usepackage{amsmath} \usepackage{mathtools} \usepackage{todonotes}
\usepackage{amssymb}  \usepackage{makeidx} \usepackage{graphicx} \graphicspath{{figures/}}
\usepackage{multicol} \usepackage[bottom]{footmisc} \usepackage[utf8]{inputenc} \usepackage{multirow}
\usepackage{booktabs} \usepackage{adjustbox}
\usepackage{caption}
\usepackage{makecell} \usepackage{subcaption}  \usepackage{tikz}
\usetikzlibrary{positioning}
\usetikzlibrary{shapes,snakes}
\usepackage{bm} \usepackage{float} \usepackage{color} \usepackage{empheq} \usepackage{threeparttable} \usepackage{xcolor}
\usepackage{xurl} \usepackage{tcolorbox}

\DeclareMathSymbol{\shortminus}{\mathbin}{AMSa}{"39}
\usepackage{paralist}

\newcommand\gb[1]{\textbf{\color{green!80!black}{#1}}}

\usepackage{mathtools}
\DeclarePairedDelimiterX{\norm}[1]{\lVert}{\rVert}{#1}

\allowdisplaybreaks
\renewcommand{\baselinestretch}{0.96}
\newcommand{\R}[2]{{}^{#2}_{#1}\mathbf{R}}

\newcommand{\angtilde}[2]{{}^{#2}_{#1}\tilde{\boldsymbol{\theta}}}

\newcommand{\p}[2]{{}^{#2}\mathbf{p}_{#1}}

\newcommand{\ptilde}[2]{{}^{#2}\tilde{\mathbf{p}}_{#1}}

\newcommand{\skw}[1]{\lfloor {#1} \rfloor}

\begin{document}

\title{\LARGE \bf
NeRF-VINS: A Real-time Neural Radiance
Field Map-based
Visual-Inertial Navigation System
}

\author{Saimouli Katragadda$^{1}$, Woosik Lee$^{1}$, Yuxiang Peng$^{1}$, Patrick Geneva$^{1}$, \\ Chuchu Chen$^{1}$, Chao Guo$^{2}$, Mingyang Li$^{2}$, and Guoquan Huang$^{1}$
\thanks{This work was partially supported by 
the University of Delaware (UD) College of Engineering, 
Delaware NASA/EPSCoR Seed Grant, 
NSF (IIS-1924897, SCH-2014264), 
and Google ARCore.
}\thanks{$^{1}$The authors are with the Robot Perception and Navigation Group (RPNG), University of Delaware, Newark, DE 19716.
Email: {\tt\small \{saimouli, woosik, yxpeng, pgeneva, ccchu, ghuang\}@udel.edu}. \newline 
${\hspace{0.65cm}}^{2}$The authors are with Google, Mountain View, CA 94043. Email: {\tt\small \{chaoguo, mingyangli\} @google.com}.}
}

\maketitle
\pagestyle{empty}

\begin{abstract}

Achieving  efficient and consistent localization with a prior  map remains challenging in robotics.
Conventional keyframe-based approaches often suffer from sub-optimal viewpoints due to limited field of view (FOV) and/or constrained motion,  thus degrading the localization performance.
To address this issue,
we design a real-time tightly-coupled Neural Radiance Fields (NeRF)-aided visual-inertial navigation system (VINS).
In particular, by effectively leveraging the NeRF's potential to synthesize novel views,
the proposed NeRF-VINS overcomes the limitations of traditional keyframe-based maps (with limited views) and optimally fuses IMU, monocular images, and synthetically rendered images within an efficient filter-based framework. 
This tightly-coupled fusion enables efficient 3D motion tracking with bounded errors. 
We extensively validate the proposed NeRF-VINS against the state-of-the-art methods that use prior map information,
and demonstrate its ability to perform real-time localization, at 15 Hz, on a resource-constrained Jetson AGX Orin embedded platform.

\end{abstract}

\section{Introduction}

The ability to achieve high-accuracy localization  is pivotal for edge devices
which have become prevalent through computation miniaturization enabling
AR/VR  \cite{ARcore,ARkit} and consumer drones \cite{Katragadda2019_AIAA,Chen2022IROS}.
The ubiquitous use of cameras and inertial measurement units (IMU) due to their low cost, low power, and small size makes the Visual-Inertial Navigation System (VINS) a critical component for the aforementioned applications \cite{Huang2019ICRA}. 
If no global information (e.g., GPS, loop-closures, or a prior map), 
VINS can only provide ego-motion tracking with ever-growing error.
Over the past two decades, a particular focus has been placed on leveraging {\em a priori} map as additional costly sensors are not required~\cite{Lynen2015_RSS,Kasyanov2017_IROS,Sarlin2019_CVPR, Geneva2022ICRA, Mourikis2009_TRO, Schneider2018_RAL, Cramariuc2022_RAL}.

A crucial component of successful map-based localization is an accurate place retrieval algorithm such as DBoW \cite{Galvez2012_TRO}, placeless \cite{Lynen2014_IC3V}, or NetVLAD \cite{Arandjelovic2016_CVPR}, which allows for recovery of correspondence information to construct constraints to historical information.
However, these methods may be vulnerable to viewpoint variations, poor viewpoint coverage limiting recall, scene ambiguities, and sensitivities to environmental changes after mapping \cite{Li2023_JR}.

To address these challenges, in this work, we propose to avoid the need for place recognition via the rendering of novel synthetic views adjacent to the current state estimate, enabling high-quality and informative loop-closure constraints that are not susceptible to these failure modes. 
Specifically, we introduce a new paradigm for map-based  localization which leverages the recent Neural Radiance Fields (NeRF)~\cite{Mildenhall2020_ECCV} advancements in deep learning to compress the collection of images, e.g. a prior keyframe image map, into a trained network, and then leverage during localization the high-fidelity image rendering of synthesize novel camera viewpoints.
While the NeRF's ability to accurately reconstruct complex environments has encouraged researchers to build dense NeRF maps \cite{Jiang2023_arxiv, Chung2023_ICRA},
we focus on achieving real-time localization on edge devices  with limited computational resources and thus look to leverage the comparably cheaper novel viewpoint rendering via hashing \cite{Mueller2022_ACM}.
To this end, we effectively leverage NeRF as an {\em a priori} map and maintain real-time drift-free VINS localization.
The main contribution of this work includes:
\begin{itemize}
\item We, for the first time, develop a real-time NeRF-VINS algorithm that fuses {\em a priori} NeRF-based map in a tightly-coupled manner to enable drift-free localization.

\item We conduct extensive numerical studies to understand the impact of different NeRF map construction methods, descriptor algorithms on rendered NeRF views, and environmental changes, thus better informing our design. \item The proposed NeRF-VINS is among the first  to demonstrate centimeter-level drift-free pose estimates on an edge platform (Jetson AGX Orin rendering at over 10 Hz) and outperform existing state-of-the-art methods.
\end{itemize}

\section{Related Work}

In this section, we provide an overview of methods related to visual and visual-inertial and NeRF-based localization.

\subsection{Prior Map-based Classical Localization}
\textit{Single-View Visual Localization:}
The classical structure-based method is the Perspective-n-Point (PnP) solver within a RANSAC loop for robustness \cite{Chum2003_BMVC, Chum2008_PAML}. 
The 2D-3D correspondences between the query image and a map points are typically found through the matching of local feature descriptors \cite{Sattler2011_ICCV, Liu2017_ICCV, Sattler2017_PAMI, Schoenberger2016_CVPR, Schoenberger2016_ECCV}.
To mitigate the complexity increase as the map size grows, image retrieval methods that narrow down the search space typically retrieve top similar matches (place recognition) and query keypoints in the region defined by these images for correspondences (local matching) \cite{Sarlin2019_CVPR, Humenberger2020_ARXIV}.
The quality of this approach heavily relies on the effectiveness of the image retrieval methods.
DBoW \cite{Galvez2012_TRO} has gained great popularity thanks to its efficiency, but recent deep learned-based HF-Net \cite{Sarlin2019_CVPR}, which leverages  NetVLAD \cite{Arandjelovic2016_CVPR} and SuperPoint \cite{Detone2018_CVPR} for global retrieval and local matching respectively, has demonstrated state-of-the-art performance in localization tasks.
Although there are end-to-end deep learning methods available, their poor accuracy and complexity still make structure-based methods appealing \cite{Kendall2015_ICCV, Kendall2017_CVPR, Yang2019_CVPR}. 
Additionally, all discussed methods can suffer from global descriptor ambiguities, particularly in scenarios with sparse images or significant changes in viewpoint, and poor recall due to limited view coverage of the scene which we aim to address through the proposed NeRF-VINS rendering paradigm.

\textit{Visual-Inertial Localization}:
As compared to single-view visual localization, visual-inertial localization aims to continuously provide estimates against a prior map and can leverage historical information to reduce the search space and thus complexity.
There is a rich literature, for which we refer the interested reader to the references in \cite{Geneva2022ICRA} for a summary.
One which is of particular relevance to this work is the open-sourced ROVIOLI \cite{Schneider2018_RAL} extension of ROVIO \cite{Bloesch2015_IROS,Bloesch2017_IJRR} which performs 2D-3D matches against an optimized global map commonly constructed using \texttt{maplab} \cite{Schneider2018_RAL,Cramariuc2022_RAL}.

\textit{SLAM Systems:}
In contrast to previous approaches that construct maps offline for accurate localization, SLAM builds maps online and utilizes them via loop closures. A typical SLAM architecture includes a real-time thread for camera pose tracking using sparse keypoints \cite{Klein2007_ISMAR}, \cite{Leutenegger2015_IJRR} or dense/semi-dense representations \cite{Engel2014_ECCV}, \cite{Forster2014_ICRA}, along with a non-real-time thread that optimizes and constructs the map. 
These methods use classical image retrieval techniques to query images for loop closure, which can be affected by limited viewpoint coverage and ambiguities.

\subsection{Neural Radiance Fields}

The work  \cite{Mildenhall2020_ECCV} introduced the NeRF methodology and revolutionized scene representation, novel view generation, and high-fidelity rendering.
Later works such as BARF \cite{Lin2021_ICCV} and NeRF  \cite{Wang2021_arxiv} have shown that knowing the exact poses is not required, while iMAP \cite{Sucar2021_ICCV} and NICE-SLAM \cite{Zhu2022_CVPR} showed that the joint optimization of poses in respect the NeRF can further improve performance.
There additionally have been works that have focused on map representation \cite{Jiang2023_arxiv}, and the integration within SLAM  \cite{Chung2023_ICRA,Rosinol2022_arxiv}.

As compared to the online generation of NeRF maps, we instead look to leverage a previously built NeRF to provide high-quality loop-closure information and bound estimator drift.
Only a few works have focused on leveraging NeRF to provide prior environmental information for the betterment of visual tracking.
iNeRF \cite{Yen2020_IROS} proposed to localize camera poses by optimizing the photometric error between the real and NeRF-generated images within a small static environment context but remained sensitive to the initial pose guess and large computational cost.
More recently, Loc-NeRF \cite{Maggio2023_ICRA} was proposed to employ a particle filter to remove the need for an initial guess.
While this method does not require any initial guess, it necessitates image rendering for each particle, which could easily become computationally prohibitive if using a large number of particles.
Another work similar spirit is by Adamkiewicz et al. \cite{Adamkiewicz2022_RAL} which leveraged a pre-trained NeRF map to localize and additionally optimize future trajectories.
As compared to these works which are constrained by rendering speed and their alignment computational complexity, the proposed NeRF-VINS combines the novel viewpoint rendering strength with the efficient, and accurate MSCKF-based VINS.

\begin{figure}[t]
    \centering
\centering
    \href{https://youtu.be/hR6R34G62Rs}{\includegraphics[trim=0mm 10mm 0mm 0mm,clip, width=\linewidth]{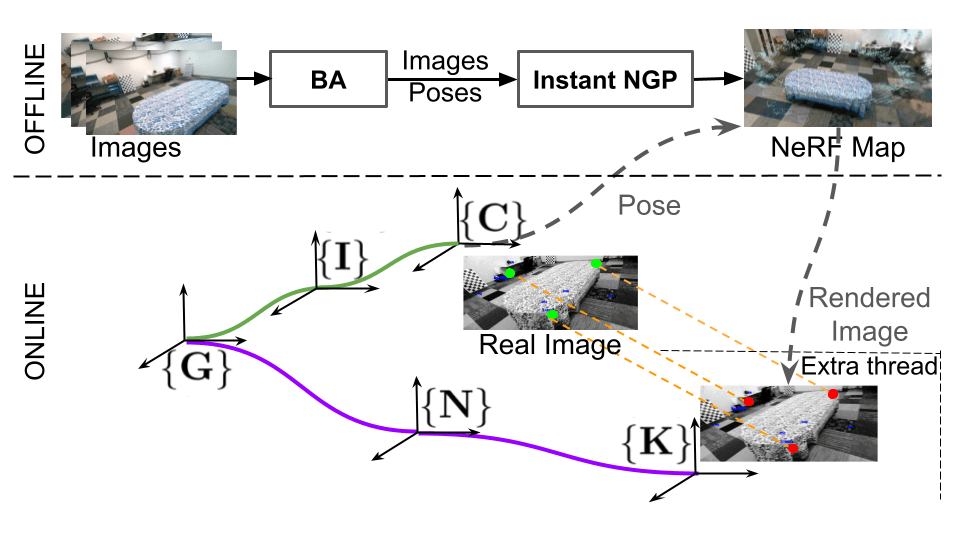}
    }
\caption{
Overview of the proposed NeRF-VINS, where $\{{G}\}$ is the global VIO frame, $\{{N}\}$ is the map frame, $\{{K}\}$ denotes the NeRF rendered image. $\{{I}\}$ and $\{{C}\}$ are IMU and camera frame, respectively. {\em Click on the image for a video demo.}
    }
    \label{fig:overview}
\end{figure}

\section{NeRF-VINS Estimator Design}

Visual-inertial localization uses two main approaches: graph optimization \cite{Leutenegger2015_IJRR,Qin2018_TRO} and filter-based methods \cite{Cramariuc2022_RAL, Mourikis2007_ICRA, Mourikis2009_TRO}. Graph optimization generally offers good accuracy, but is computationally demanding. 
In contrast, filter-based approaches like the MSCKF \cite{Mourikis2007_ICRA} efficiently integrate camera and IMU measurements, suitable for real-time applications.
The MSCKF balances feature efficiency, avoiding complexity growth, making it ideal for resource-constrained real-time localization.
The proposed NeRF-VINS estimator extends the MSCKF \cite{Mourikis2007_ICRA,Geneva2020ICRA} to fuse the prior NeRF map in a tightly-coupled manner
(see Fig.~\ref{fig:overview}).
As such, for presentation brevity, in the following, we will primarily focus on visual measurement update.

In particular, at time $t_{k}$, the system state $\mathbf{x}_k$ consists of the current inertial navigation states $\mathbf{x}_{I_k}$, historical IMU poses $\mathbf{x}_{T_k}$, and a subset of 3D environmental point features, $\mathbf{x}_f$:
\begin{align}
    \mathbf{x}_k &= 
    \begin{bmatrix}
    \mathbf{x}_{I_k}^{\top} ~
    \mathbf{x}_{T_k}^{\top} ~
    \mathbf{x}_{f}^{\top}
    \end{bmatrix}^{\top} \label{eq:state}
    \begin{aligned}
    ,\mathbf{x}_f &=
    \begin{bmatrix}
    {}^{G}\mathbf{p}_{f_1}^{\top} ~
    \dots ~
    {}^{G}\mathbf{p}_{f_i}^{\top}
    \end{bmatrix}^{\top}
    \end{aligned}
    \\
\mathbf{x}_{I_k}
    &=
    \begin{bmatrix}
    \label{eq:state_imu}
    {}_{G}^{I_k}\Bar{q} ^{\top} ~
    {}^{G}\mathbf{p}_{I_k} ^{\top} ~
    {}^{G}\mathbf{v}_{I_k} ^{\top} ~~
    \mathbf{b}_{g} ^{\top}~~
    \mathbf{b}_{a} ^{\top} 
    \end{bmatrix}^{\top} 
\\
\mathbf{x}_{T_k}
    &=
    \begin{bmatrix}
    {}_{G}^{I_{k}}\Bar{q} ^{\top} ~
    {}^{G}\mathbf{p}_{I_{k}} ^{\top} ~
    \dots ~
    {}_{G}^{I_{k-c}}\Bar{q} ^{\top} ~
    {}^{G}\mathbf{p}_{I_{k-c}} ^{\top}
    \end{bmatrix}^{\top}
\end{align}
where $ {}_{G}^{I}\Bar{q}$ is the unit quaternion (${}_{G}^{I}\mathbf{R}$ in rotation matrix form) that represents the rotation from the global $\{G\}$ to IMU frame $\{I\}$. ${}^{G}\mathbf{p}_{I}$, ${}^{G}\mathbf{v}_{I}$, and ${}^{G}\mathbf{p}_{f_i}$ are the IMU position, velocity, and $i$'th point feature position in $\{G\}$;
$\mathbf{b}_{g}$ and $\mathbf{b}_a$ are the gyroscope and accelerometer biases.
{Note that other state variables can be included, e.g., spatial-temporal calibration, but have been omitted for clarity.}

The state is propagated over time based on the IMU measurements.
A canonical three-axis IMU provides linear acceleration, ${}^I\mathbf a_m$, and angular velocity measurements, ${}^I\bm\omega_m$. 
The IMU nonlinear kinematics is generically given by \cite{Chatfield1997}:
\begin{align}
    \mathbf{x}_{I_{k+1}} 
    &
    = \mathbf{f}\left(\mathbf{x}_{I_k}, {}^{I}\mathbf a_{k}, {}^{I}\boldsymbol\omega_{k}, \mathbf{n}_{I_k}\right)
    \label{eq:imu_eq}
\end{align}
where
$\mathbf{n}_{I_k} = [\mathbf{n}^{\top}_{g}~\mathbf{n}^{\top}_a~\mathbf{n}^{\top}_{wg}~\mathbf{n}^{\top}_{wa}]^{\top}$; $\mathbf{n}_{g}$ and $\mathbf{n}_a$ are Gaussian white noises, and $\mathbf{n}_{wg}$ and $\mathbf{n}_{wa}$ are the random walk bias noises of gyroscope and accelerometer, respectively.
With this model~\eqref{eq:imu_eq}, we can perform EKF propagation of the state estimate and covariance~\cite{Mourikis2007_ICRA}.

\subsection{Measurement Update with Real Images}
\label{sec:feat}

As in \cite{Geneva2020ICRA}, bearing measurements of detected features seen at time $t_k$ are modeled as follows:
\begin{gather}
\label{eq:zC}
\mathbf{z}_{C_k} = \mathbf{h}_c(\mathbf{x}_{T_k},{}^{G}\mathbf{p}_f) + \mathbf{n}_{C_k}
:= \boldsymbol{\Lambda}({}^{C_k}\mathbf{p}_f) + \mathbf{n}_{C_k} \\{}^{C_k}\mathbf{p}_f =
{}^{C}_{I}\mathbf{R}~{}^{I_k}_{G}\mathbf{R}({}^{G}\mathbf{p}_f-{}^{G}\mathbf{p}_{I_k}) + {}^{C}\mathbf{p}_I 
\end{gather}
where $\boldsymbol{\Lambda}\left([x~y~z]^\top\right)=\begin{bmatrix} x/z & y/z \end{bmatrix}^\top$ and $\mathbf{n}_{C_k}$ is the white Gaussian  noise.
Linearizing  Eq. \eqref{eq:zC} yields the following measurement residual:
\begin{align}
\mathbf r_{C_k}
&= \mathbf z_{C_k} - \mathbf{h}_c(\hat{\mathbf{x}}_{T_k}, {}^G\hat{\mathbf{p}}_{f} ) \\
&\simeq \mathbf{H}_{T_k}\tilde{\mathbf{x}}_{T_k} + \mathbf{H}_{f_k} {}^G\tilde{\mathbf{p}}_f + \mathbf{n}_{C_k} \label{eq:real_uv_lin}
\end{align}
where $\mathbf{H}_{T}$ and $\mathbf{H}_{f}$ are the Jacobian matrix of the measurement with respect to each state.
We keep the long-tracked features in the state till lost in order to leverage their future observations, 
while the short-tracked features are updated via the efficient MSCKF nullspace projection \cite{Mourikis2007_ICRA}.

\subsection{Measurement Update with NeRF Images}

When a camera image reading is received, a NeRF render is triggered at a pose with a small horizontal positional offset (e.g., 10 cm, as in our experiments, in analogy to a stereo baseline) to the current camera pose.
This synthetic image should have a \textit{significant} overlapping field of view (FOV) with the current real image, 
which facilitates high-quality feature matching.
The small positional offset also enables robust triangulation and accurate feature matching between the real and synthetic images even when the camera is static.

Once the rendering is completed, descriptor-based feature matching is performed to the current image, where a 2D-to-2D prior keyframe measurement model is leveraged \cite{Geneva2019ICRA}.
For example, consider that from the rendered image we get a bearing measurement, $\mathbf{z}_{N_k}$, which is related by [see \eqref{eq:zC}]:
\begin{align}
\mathbf{z}_{N_k} &= \mathbf{h}_n\left({}^{G}\mathbf{p}_{f}\right) + \mathbf{n}_{N_k}
:=  \boldsymbol{\Lambda}({}^{K}\mathbf{p}_f) + \mathbf{n}_{N_k} \\\label{eq:nerf_up}
{}^{K}\mathbf{p}_f &= \p{N}{K} + s\R{N}{K}(\p{G}{N} + \R{G}{N}\p{f}{G})
\end{align}
where $s$ is the scale factor of the map and $\mathbf{n}_{N_k}$ is the zero mean Gaussian noise.
Note that we model the bearing as only a function of the feature $\p{f}{G}$, and consider the map transform $\{s, \R{G}{N}, \p{G}{N}\}$ to be known (see Sec. \ref{ch:map_align}) and the rendered camera pose $\{\R{N}{K}, \p{N}{K}\}$ to have some known orientation and position error $\{\angtilde{G}{N}, ~ \ptilde{G}{N}\}$.
Thus, we have the following linearized model:
\begin{align}
    \mathbf{r}_{N_k} 
    &= \mathbf{z}_{N_k} - \mathbf{h}_n({}^{G}\hat{\mathbf{p}}_{f})
= 
    s \mathbf{H}_{\boldsymbol{\Lambda}} \R{N}{K} \R{G}{N} \ptilde{f}{G} + \mathbf{n}_{N_k}'   \\
\mathbf{n}_{N_k}' &= s \mathbf{H}_{\boldsymbol{\Lambda}} \R{N}{K} (\skw{\R{G}{N}\p{f}{G} \times} \angtilde{G}{N} + \ptilde{G}{N}) + \mathbf{n}_{N_k}
\end{align}
where $\mathbf{H}_{\boldsymbol{\Lambda}}$ is the measurement jacobian in respect to the 3D point feature and $\skw{\cdot\times}$ is the skew-symmetric matrix.
The linearized model can be  used to update the features in the state or can be stacked with the real image measurements~\eqref{eq:real_uv_lin} to perform   (SLAM or MSCKF) EKF update.

\section{System Integration}

\begin{figure}[t]
    \centering
    \includegraphics[trim=0mm 0mm 0mm 14mm,clip, width=0.46\columnwidth]{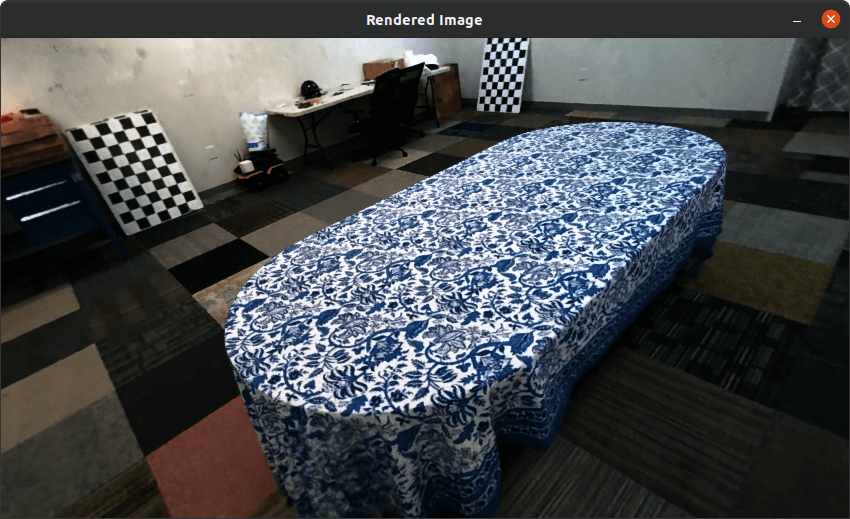}
    \includegraphics[trim=0mm 0mm 0mm 14mm,clip, width=0.46\columnwidth]{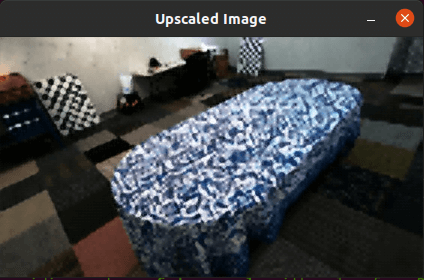}
    \caption{Example rendered images for testing matching methods. \textit{Left}: Rendered image with resolution 424$\times$240.
    \textit{Right}: Rendered image with 141$\times$80 resolution
    and up-scaled to 424$\times$240 with FSRCNN \cite{Dong2016_ECCV}.
    }
    \label{fig:downsample3_example}
\end{figure}

\begin{figure*}[t]
\centering
\begin{minipage}{0.94\textwidth}
    \centering
\includegraphics[trim=10mm 0mm 10mm 11mm,clip, width=0.98\linewidth]{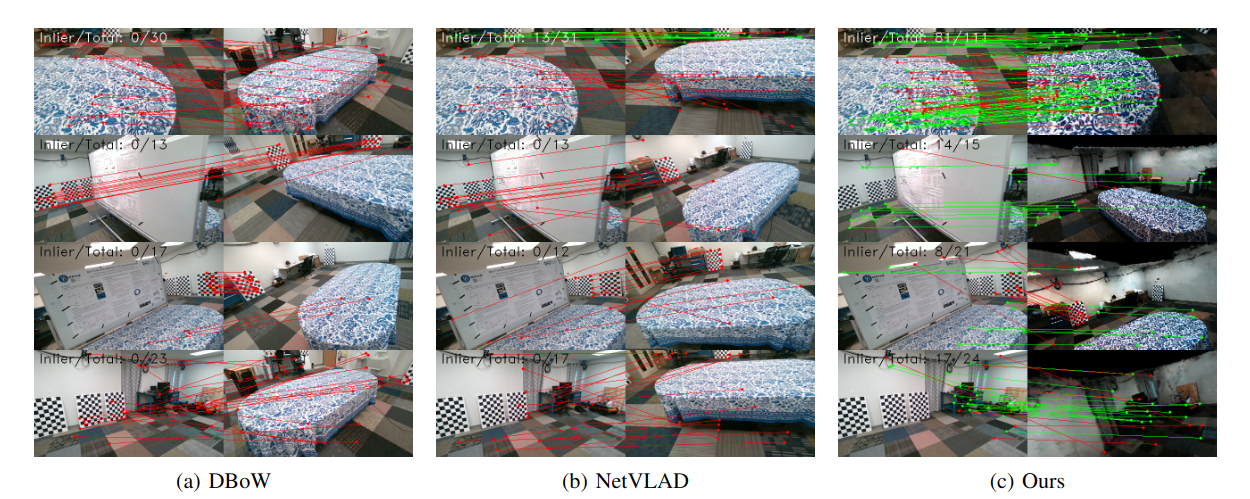}
\end{minipage}
\caption{Qualitative study of failure cases of classical place recognition method. Green and Red lines indicate inliers and outliers, respectively. Input image (left of each column) and retrieved, rendered for the NeRF case (resolution 212$\times$140 and upsampled to 424$\times$240), image is shown (right of each column). Images are shown in color for visualization purposes.
}
\label{fig:failure_case}
\end{figure*}

\begin{figure}\centering
    \includegraphics[trim=8mm 8mm 8mm 8mm,clip,width=0.99\linewidth]{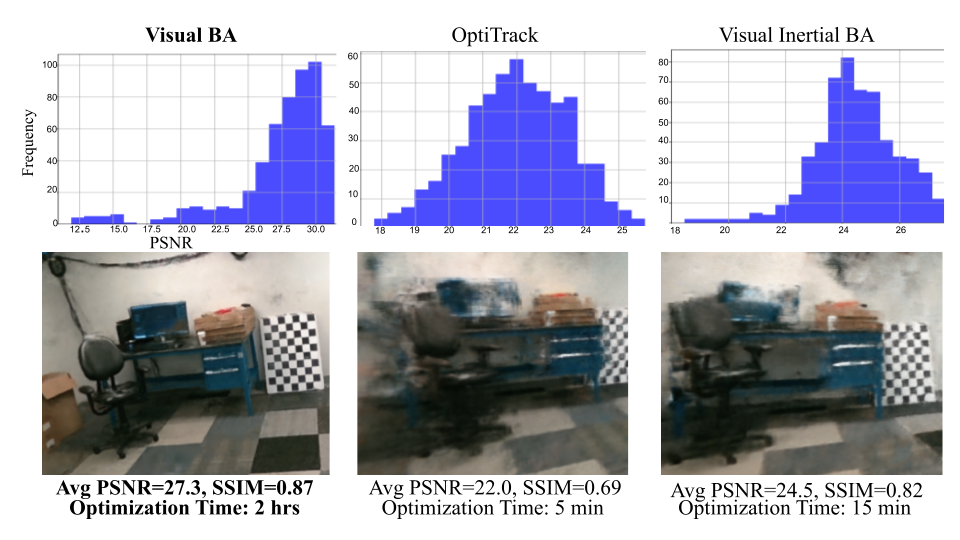}
    \caption{Qualitative comparison of NeRF Map trained with different methods using 543 keyframe images. The top row shows the PSNR histograms and the bottom row shows exemplary images rendered from each method.}
\label{fig:psnr_hist} 
\end{figure}

Armed with the NeRF-VINS estimation theory presented in the previous section, 
we now describe how to integrate the NeRF model and feature matching between synthetic and real images to form a tightly-coupled system.

In particular, our  system leverages the open-source InstantNGP \cite{Mueller2022_ACM} for rendering and prior map training.
The OpenVINS \cite{Geneva2020ICRA} frontend is modified to incorporate SuperPoint descriptors using Tensor-RT pipeline \cite{TensorRT}. 
We used OpenCV \cite{Bradski2000} and CUDA to convert GPU-rendered images to a 32bit-float RGB image on the CPU. 
Additional care has been taken to convert the NeRF-rendered image coordinate system to a right-hand coordinate system by inverting the y and z axes of InstantNGP.
The code is written in C++ and CUDA and runs on Jetson AGX Orin unless specified.

\subsection{Feature Descriptor Selection} \label{sec:impl_desc}

A crucial component is the ability to match features between the current frame and the rendered NeRF viewpoint.
Thus significant effort has been spent to investigate the performance of various feature matching methods such as AKAZE \cite{alcantarilla2011fast}, KAZE \cite{alcantarilla2012kaze}, BRISK \cite{leutenegger2011brisk}, ORB \cite{rublee2011orb}, and the selected SuperPoint (SP) \cite{Detone2018_CVPR}.
For this test, we choose a challenging scenario by rendering at (141$\times$80) and upscaling to 424$\times$240 using FSRCNN \cite{Dong2016_ECCV}, see Fig. \ref{fig:downsample3_example}.
Shown in Tab. \ref{tab:descriptor_test}, the average descriptor extraction time, number of matches between the rendered and current camera image, and Absolute Trajectory Error (ATE) \cite{Zhang2018_IROS} of VINS for each method have been compared.
It is clear that the handcrafted matching methods (AKAZE, BRISK, ORB, and KAZE) often fail and show large errors which is expected due to the limited fidelity in the up-sampled resolution NeRF image.
On the other hand, SuperPoint (SP) and its optimized variant (SP Opt.) are shown to be robust to these conditions and report the highest accuracy and shortest descriptor extraction time. 
We thus select the optimized SuperPoint for its robustness and efficiency for synthetic NeRF to real image matching.

\subsection{Image Rendering and Feature Matching}\label{sec:jetson_opt}

\begin{table}[]
\centering
\captionsetup{skip=0.1ex} \caption{Average descriptor extraction time, number of matches, and ATE reported on the UD AR Table 1-8 dataset \cite{Chen2023ICRA} for different matching methods utilizing the configuration depicted on the right side of Fig. \ref{fig:downsample3_example}.}
\begin{adjustbox}{max width=1\columnwidth,center}
\setlength{\tabcolsep}{4pt}
\label{tab:descriptor_test}
\begin{tabular}{ccccccccccccc} 
\toprule
 & \textbf{AKAZE}  & \textbf{BRISK} & \textbf{ORB} & \textbf{KAZE} & \textbf{SP} & \textbf{SP Opt.}\\\midrule
Time (ms) & 31 & 88 & 13 & 140 & 15 & 7 \\
No. of Matches & 55 & 85 & 20 & 117 & 31 & 30 \\
ATE (deg/m) & 2 FAIL & 5 FAIL  & 6 FAIL & 2.40 / 0.29 & 1.16 / 0.15 & 1.18 / 0.16\\
\bottomrule
\end{tabular}
\end{adjustbox}
\end{table}

Rendering NeRF images remains a computationally expensive operation even with state-of-the-art techniques \cite{Mueller2022_ACM}.
On embedded devices like the Jetson AGX Orin, it takes approximately 660ms (2Hz) to render an image with dimensions 424$\times$240.
To improve render speed and minimize loop-closure latency, we use a two-step process. Initially, we generate NeRF renders at half resolution (212$\times$140).
Then, we employ the lightweight FSRCNN \cite{Dong2016_ECCV} for up-sampling to the original size of 424$\times$240.
This approach strikes a balance between computational speed and image quality (see Fig. \ref{fig:downsample3_example} and Tab. \ref{tab:descriptor_test} for the extreme case of 141$\times$80 resolution). 
We further reduce the resolution levels and the hashing size of the model in InstantNGP \cite{Mueller2022_ACM}
and minimize multiple CPU copies by directly transferring rendered images to our localization pipeline for descriptor extraction.

The rendering is run on a separate thread to prevent blocking of the real-time VINS.
The SuperPoint feature matching network has been modified to use a lightweight ResNet18 \cite{He2016_CVPR} and optimized to support a 16-bit floating point using TensorRT \cite{TensorRT}.
This secondary thread, which performs rendering and matching, runs at 15Hz on the Jetson.
Additional timing details are reported in \cite{Katragadda2023TRNERF}.

\subsection{Offline NeRF  Map Generation} \label{ch:map_align}

Another foundational component is the ability to build and train a prior NeRF map which can be leveraged online (see Fig.~\ref{fig:overview} top half).
The first challenge is to recover accurate camera poses which can then be used in conjunction with images to train the NeRF model.
Three different methods were investigated: (i) Visual Bundle Adjustment (BA) via COLMAP \cite{Schoenberger2016_CVPR, Schoenberger2016_ECCV}, (ii) Visual-Inertial BA via \texttt{maplab} \cite{Cramariuc2022_RAL}, and (iii) fusion of OptiTrack poses with IMU via \textit{vicon2gt} \cite{Geneva2020TRVICON2GT}.
We leveraged the keyframing selection in \texttt{maplab} to select a subset of 543 of poses which both COLMAP and \texttt{maplab} optimized.

Analyzing the results of the Table 5 dataset in Fig. \ref{fig:psnr_hist}, we observe clear variation in Peak Signal-to-Noise Ratio (PSNR) and Structural Similarity Index (SSIM) \cite{Mildenhall2020_ECCV}. 
COLMAP's up-to-scale Visual BA yields superior values, albeit at the expense of significant computation to optimize the camera poses.
Conversely, the Visual-Inertial BA in \texttt{maplab} did take less time to optimize, but suffers in PSNR whose blurriness can be seen in the exemplary images.
A similar trend is observed in the \textit{vicon2gt} OptiTrack+IMU results, indicating that while the fusion of inertial information accelerates optimization time and provides scale information, it does not improve visual reconstruction quality, likely due to calibration and sensor synchronization errors in this dataset. 
We additionally conjecture that IMU-aided methods likely do not fully minimize visual reprojection errors to the same degree as COLMAP, potentially leading to suboptimal poses for desired geometric reprojection errors crucial for high-quality NeRF creation.
We thus opted to use the up-to-scale COLMAP poses for training the NeRF.
These poses were aligned to the groundtruth poses based on similarity transformation (sim3) to remove the scale ambiguity.
For each dataset we assume that the proposed NeRF-VINS has been initialized in the NeRF prior map and directly leverage a pre-computed map transform.

 \section{Experimental Validation}

We validate the proposed NeRF-VINS and baseline methods on the recently released AR Table Dataset \cite{Chen2023ICRA}.
This dataset is ideal for NeRF reconstruction due to its object-centric trajectories which observe a table placed centrally.
This dataset additionally enables us to evaluate the robustness of algorithms to changing environments (see  Fig. \ref{fig:table_dataset}), 
due to the addition of a whiteboard for the three datasets (Table 5-7) and the moving of the table to the side of the room in Table~8.
Unless specifically noted, all prior map methods leverage Table 1 for datasets 1-4 and Table 5 for 5-7.

In particular,  for comprehensive validation, we  evaluated the following state-of-the-art methods:

\subsubsection{\textbf{Single-Shot Visual Localization}} The open-source Hierarchical Localization (HLoc) system \cite{Sarlin2019_CVPR} that used NetVLAD for image retrieval, and SuperPoint \cite{Detone2018_CVPR} descriptor establishes a baseline for expected state-of-the-art performance.
In this system, local matching is performed using a nearest-neighbor search with a ratio test and geometric verification, which aligns with our pipeline. Notably, the use of Lightglue \cite{Lindenberger2023_ICCV} matching remains computationally expensive (16 ms for a pair, thus 800 ms for top 50 on A3000 GPU) and did not yield substantially better results in the evaluated dataset.
The same images and poses that are used to train the NeRF are leveraged in its map.
We evaluated the performance with the top 5 and 50 nearest neighbor matches: HLoc (top5) and HLoc (top50), respectively.
Due to its single-shot nature, we found that for many image localization accuracy was poor, and thus in most results presented we select an inlier set of good quality success to provide a reasonable comparison.
Note that this contrasts the below map-based methods and proposed NeRF-VINS which provide \textit{continuous} estimates.

\subsubsection{\textbf{Map-based Visual-Inertial Localization}} 
For map-based VINS, the filter-based ROVIO with additional re-localization module \cite{Bloesch2017_IJRR} (ROVIOLI) from \texttt{maplab} \cite{Cramariuc2022_RAL} provides one of the closest direct comparisons to the proposed method.
We report the accuracy of both the odometry, ROVIOLI, and the map-aided, ROVIOLI+Map, which leverages the \texttt{maplab} optimized prior map with the same keyframes used to train the NeRF.
VINS-Fusion (VF) \cite{qin2019general}, is additionally compared against as it has support to re-localization against a previous-built relative pose graph using DBoW2 \cite{Galvez2012_TRO}.
Thus we run VF on the prior map dataset to generate a pose graph that is then leveraged for sequential datasets (e.g. the whole dataset Table 1 is processed, as compared to the proposed which uses only a small subset of keyframes).
Both the odometry, VF, the secondary pose graph without relocalization, VF+Loop, and then the secondary thread which is able to relocalize against the prior map pose graph, VF+Loop+Map, are evaluated.

\subsection{Localization Accuracy}

\begin{table*}[t]
\centering
\caption{
The ATE of each 
algorithm on the AR Table dataset (degree/cm). The top two best results are highlighted with a bold green color.
}
\newcommand{\tmt}[2]{\multirow{#1}{*}{\rotatebox[origin=c]{90}{#2}}}
\label{tab:ate_table_res}
\begin{adjustbox}{max width=\textwidth,center}
\begin{threeparttable}
\begin{tabular}{cc|ccccccccccc}\toprule
\multicolumn{2}{c|}{\textbf{Algorithms}}   & \textbf{Table 1} & \textbf{Table 2} & \textbf{Table 3} & \textbf{Table 4} & \textbf{Table 5}  & \textbf{Table 6}  & \textbf{Table 7} & \textbf{Average} \\\midrule
\tmt{6}{\textbf{Map-based}}  
& Nerf-VINS (D) & 0.51 / 1.8 & 0.27 / 1.0 & 0.50 / 1.0 & 0.35 / 1.5 & 0.43 / 1.4 & 0.59 / 1.9 & 0.46 / 1.6 & \gb{0.44} / \gb{1.5} \\ 
& Nerf-VINS (J) & 0.47 / 2.0 & 0.29 / 0.8 & 0.50 / 0.9 & 0.31 / 1.6 & 0.43 / 1.3 & 0.54 / 1.9 & 0.51 / 1.7 & \gb{0.44} / \gb{1.5} \\ & VF+Loop+Map           & 0.93 / 4.1 & 1.27 / 7.1 & 0.88 / 6.1 & 1.39 / 5.2 & 0.72 / 3.2 & 0.93 / 3.7 & 1.68 / 5.3 & 1.11 / 5.0 \\ & ROVIOLI+Map           & 0.54 / 2.1 & 1.30 / 3.6 & 0.67 / 2.2 & 1.15 / 4.3 & 0.86 / 3.7 & 2.33 / 17.9& 2.42 / 13.6 & 1.32 / 6.8 \\ & HLoc (top5)*           & 0.41 / 1.0 & 0.40 / 1.6 & 0.38 / 1.4 & 0.31 / 1.3 & 0.41 / 1.2 & 0.60 / 1.6 & 0.51 / 2.0 & 0.48 / 1.4 \\ & HLoc (top50)*          & 0.41 / 1.0 & 0.33 / 1.4 & 0.35 / 1.2 & 0.30 / 1.2 & 0.40 / 1.2 & 0.57 / 1.6 & 0.51 / 2.0 & \gb{0.45} / \gb{1.3} \\ \midrule
\tmt{4}{\textbf{VINS}}  
& OpenVINS              & 1.17 / 5.4 & 0.55 / 2.2 & 1.02 / 3.4 & 1.21 / 5.9 & 0.50 / 3.3 & 1.04 / 3.7 & 1.31 / 7.2 & 0.97 / 4.5 \\
& ROVIOLI               & 2.05 / 7.1 & 1.11 / 4.1 & 2.63 / 7.9 & 1.48 / 11.1 & 2.50 / 12.1 & 1.10 / 4.3 & 3.12 / 15.9 & 2.00 / 8.9 \\ & VF+Loop               & 1.25 / 6.7 & 1.18 / 9.2 & 0.95 / 6.5 & 1.10 / 5.7 & 0.88 / 2.8 & 0.98 / 11.2 & 1.57 / 10.1 & 1.13 / 7.5 \\
& VF                    & 1.62 / 5.8 & 1.32 / 3.0 & 1.47 / 7.6 & 1.75 / 5.6 & 1.12 / 3.4 & 0.98 / 5.3 & 1.67 / 9.3 & 1.42 / 5.7 \\
\arrayrulecolor{black}\bottomrule
\end{tabular}
\begin{tablenotes} \footnotesize
\item[*] Large failures (errors larger than 5 degrees or 10 cm) of HLoc (top5) and HLoc (top50) are excluded from statistics: \\
\hfill\hfill HLoc (top5) failure rates: Table 2 37$\%$, Table 3 5.5$\%$, Table 4 0.4$\%$, Table 5 0.5$\%$, Table 6 1$\%$, Table 7 0.5$\%$\\
\hfill\hfill HLoc (top50) failure rates: Table 2 39$\%$, Table 3 2.4$\%$, Table 4 0.4$\%$
\end{tablenotes}
\end{threeparttable}
\end{adjustbox}
\end{table*}
 
\begin{figure}\centering
    \includegraphics[trim=0mm 0mm 0mm 14mm,clip, width=0.96\linewidth]{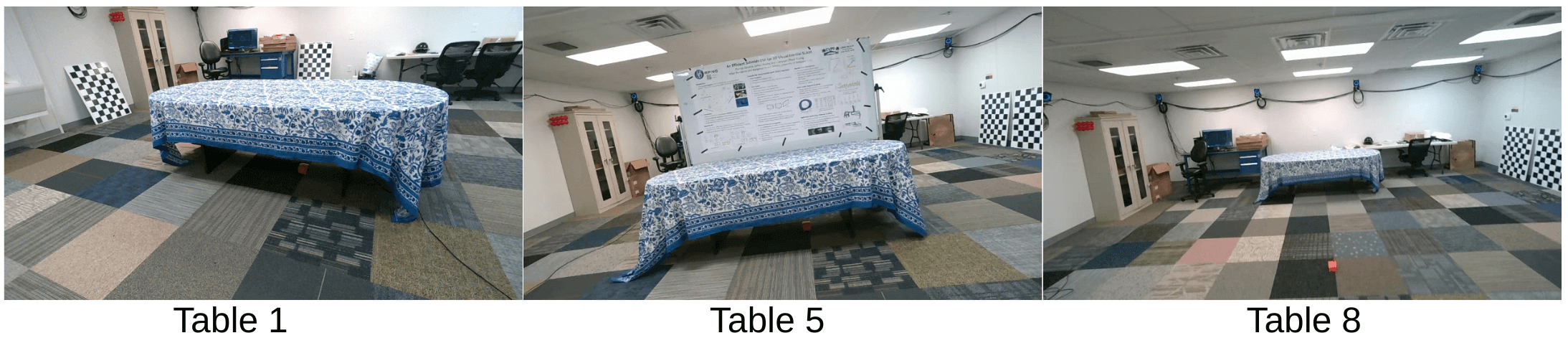}
    \caption{
Exemplary environment configurations in \cite{Chen2023ICRA}.}
    \label{fig:table_dataset}
\end{figure}

\begin{figure}
    \centering
    \includegraphics[trim=0mm 0mm 0mm 1mm,clip, width=0.99\linewidth]{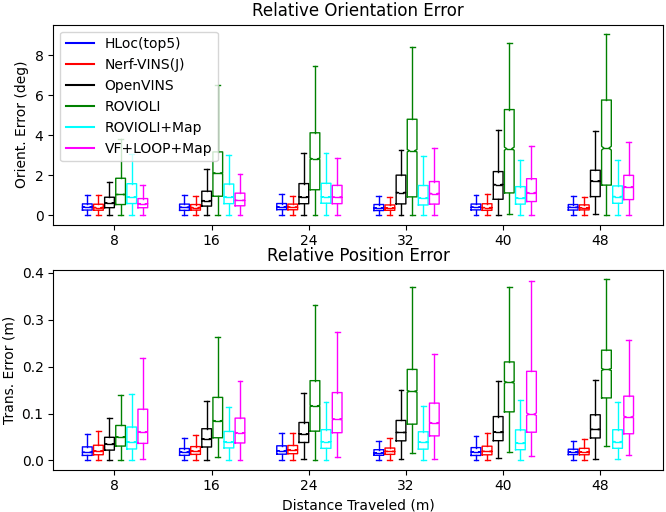}
    \caption{
    Boxplot of the RPE \cite{Zhang2018_IROS} statistics run with the same setting as Tab. \ref{tab:ate_table_res}. The box spans the first and third quartiles, while the whiskers are the upper and lower limits.
    }
    \label{fig:rpe_plot}
\end{figure}

Tab. \ref{tab:ate_table_res} shows the ATE of all methods including the proposed method on our desktop (termed Nerf-VINS (D)) equipped with an A4500 NVIDIA graphics card and on Jetson AGX Orin (termed Nerf-VINS (J)), we also provide odometry methods as a reference to show that we are able to improve the system we build on (i.e OpenVINS). 
It is clear that our proposed method achieved one of the best accuracy over all algorithms while HLoc showed competing results (note we excluded large failures of HLoc from statistics). 
An interesting observation is that VF reported higher accuracy than VF+Loop which was due to multiple false loop closures induced by incorrect DBoW matching.
This poor performance is shown by other methods which leverage DBoW, showing the need for novel view synthesis.

The Relative Pose Error (RPE) \cite{Zhang2018_IROS}  over all datasets (see Fig. \ref{fig:rpe_plot}) highlights the significant advantage of incorporating NeRF map features, which effectively mitigates drift and maintains bounded error. 
We attribute this performance gain
to the proposed method's ability to render informative novel scenes resulting in good viewpoints and a good number of quality measurements (Fig. \ref{fig:failure_case}).
Though HLoc was able to provide good accuracy, there were many failures that were excluded from the statistics, and moreover, the classification of inliers and outliers for real-time estimation is challenging. 

\begin{table}[t]
\centering
\caption{Average timing for proposed NeRF-VINS and HLoc pipeline in milliseconds. Recorded on a laptop with A3000 GPU and 11th Gen Intel(R) Core(TM) i7-11800H @ 2.30GHz CPU.}
\begin{adjustbox}{max width=\columnwidth,center}
\begin{tabular}{ccccccccccccc}
\toprule
\textbf{Step}& \textbf{Nerf-VINS (D)}  & \textbf{HLoc (top 5)} & \textbf{HLoc (top 50)} \\\midrule
Tracking & 8.5 & - & -\\
Rendering / NetVLAD & 11.6 & 12.9 & 12.9  \\
Superpoint Extraction & 5.4 & 7.6 & 7.6 \\
Local Matching & 1.7 & 15.2 & 153.7 \\
Update / PnP & 2.5 & 21.3 & 157.7 \\
Total  & 29.8  & 57.0 & 331.9\\
\bottomrule
\label{tab:hloc_time}
\end{tabular}
\end{adjustbox}
\end{table}

\subsection{Computational Complexity}

We additionally investigated the average timing of each function of our system and compared it with HLoc.
Note that we disabled the multi-threading of the proposed NeRF-VINS and compare on the same system for a fair comparison.
The results reported in Tab. \ref{tab:hloc_time} show the total time of the proposed system takes 30 ms which is almost half of the total timing of HLoc with top 5 match results. 
Though the performance of HLoc can be improved by retrieving more images, this will introduce a significant computation burden for local matching and PnP, making it difficult to run in real-time (HLoc (top 50) pipelines take 331.9 ms per frame as shown in Tab. \ref{tab:hloc_time}).
This clearly shows that our pipeline is lightweight and is capable of high-rate rendering of the NeRF images enabling real-time localization fully leveraging the NeRF map information.

\begin{table}
\centering
\caption{AR table ATE (degree/cm) and Table 1 is used as a map for the following sequence. Blanks indicate failures. The top two best results are indicated with bold green color.}
\label{tab:ate_table_env_change}
\begin{adjustbox}{max width=\columnwidth,center}
\begin{threeparttable}
\setlength{\tabcolsep}{2.5pt}
\begin{tabular}{cccccccc}\toprule
\textbf{Algorithm} & \textbf{Table 5}  & \textbf{Table 6}  & \textbf{Table 7}  & \textbf{Table 8} & \textbf{Average} \\\midrule
Nerf-VINS (J)   & 0.49 / 3.0 & 0.61 / 4.1 &   0.54 / 3.3 & 0.38 / 3.0 & \gb{0.50} / \gb{3.4} \\
HLoc (top5)*    & 0.61 / 3.5 & 0.64 / 3.6 &   0.61 / 3.1 & 0.50 / 3.7 & \gb{0.59} / 3.5  \\
HLoc (top50)*   & 0.65 / 3.4 & 0.67 / 3.6 &   0.62 / 3.0 & 0.47 / 3.1 & 0.60 / \gb{3.3}  \\
VF+Loop+Map     & 0.95 / 12.4 & 0.82 / 3.3 & 1.60 / 9.3 & 2.44 / 9.9 & 1.45 / 8.7 \\
ROVIOLI+Map     & 2.48 / 11.3 & 1.89 / 12.9 & 2.59 / 14.8 & - / -    & 2.32 / 13.0 \\
\arrayrulecolor{black}\bottomrule
\end{tabular}
\begin{tablenotes} \footnotesize
\item[*] HLoc error larger than 5 degrees or 10 cm are removed to be presentable \\
\hfill\hfill HLoc(top5) failure rates: Table 5 38.9$\%$, Table 6 36.8$\%$, Table 7 37.1$\%$, Table 8 30.5$\%$ \\
\hfill\hfill HLoc(top50) failure rates: Table 5 23.9$\%$, Table 6 29.4$\%$, Table 7 21.8$\%$, Table 8 10.7$\%$
\end{tablenotes}
\end{threeparttable}
\end{adjustbox}
\end{table}

\subsection{Robustness to Environment Changes}

To assess our system in generating favorable viewpoints enabling robust localization even when the environment is changed after mapping, we examined a more challenging scenario: employing Table 1 as the map and running on Table 5-8 each with distinct environments (refer to Fig. \ref{fig:table_dataset}).
Our system shows robust localization performance which is also competitive with HLoc (note that HLoc encounters numerous failures, which are omitted from consideration see Tab. \ref{tab:ate_table_env_change}).
In contrast, our system consistently delivers advantageous viewpoints, facilitating large inlier measurements (Fig. \ref{fig:failure_case}).

\begin{figure}[t]
    \centering
\includegraphics[trim=4mm 4mm 1mm 7mm,clip, width=0.99\columnwidth]{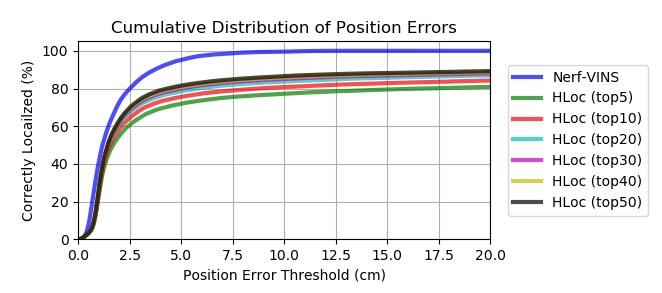}
    \caption{The percentage of images successfully localized under a certain position error threshold using Table 1 as a map to evaluate Table 1-8. }
    \label{fig:recall_plot}
\end{figure}

As can be seen from Fig. \ref{fig:recall_plot}, around 80\% percent of images for our pipeline are localized within a 2.5 cm accuracy threshold, while HLoc is only around 70\% when matching with the top 50 images.
Our system can localize almost all the images within a 7.5 cm position error, while HLoc using the top 5 images and top 50 can only localize 80.9\% and 89.3\%  images within a 20 cm error bound, respectively.

\subsection{Discussion and Limitations}
We observe that rendering images at adjacent poses generates more matches between the rendered image and the camera image compared to queried database images like those in DBoW. This suggests that the rendered image, being pose-based, is less influenced by scene ambiguities, particularly noticeable during environmental changes in Fig. \ref{fig:failure_case}.
The results presented in Fig. \ref{fig:psnr_hist} raise intriguing considerations and challenges regarding the necessity to expedite training time while preserving rendering quality in joint optimization. 
Additionally, studying the sensitivity of IMU noise and its effects on rendering quality and joint optimization costs warrants further in-depth investigation.
While we have demonstrated that the proposed method exhibits superior localization performance, similar to other NeRF methods, our map is also object-centric.
To train the map effectively, requires surrounding images for effective training. One potential solution is to leverage F2-NerF \cite{Wang2023_CVPR} and Block-Nerf \cite{Tancik2022_CVPR}, designed for unbounded camera trajectories and large-scale map training. Recent works such as Kerbl et al. \cite{Kerbl2023_ACM} offer greater rendering speed and open a new avenue for exploration.
We leave these as future work.

\section{Conclusions and Future Work}

In this paper, we have developed a real-time tightly-coupled NeRF-VINS algorithm.
Built on top of the MSCKF, the proposed NeRF-VINS extends to efficiently and accurately fuses the NeRF synthetic images 
to overcome the limited viewpoint challenges commonly encountered by the keyframe map-based localization methods. 
In particular, as NeRF can generate novel views from any viewpoint, 
we exploit this advantage to synthesize  better views to provide higher inlier matches that allow for full utilization of the map information, resulting in performance gain.
In the future, we will investigate NeRF map-based initialization i.e., initializing the transform between the IMU and map frames.

{
\newpage
\bibliographystyle{packages/IEEEtran}
\bibliography{libraries/extra,libraries/related,libraries/rpng}

\begin{thebibliography}{10}
\providecommand{\url}[1]{#1}
\csname url@rmstyle\endcsname
\providecommand{\newblock}{\relax}
\providecommand{\bibinfo}[2]{#2}
\providecommand\BIBentrySTDinterwordspacing{\spaceskip=0pt\relax}
\providecommand\BIBentryALTinterwordstretchfactor{4}
\providecommand\BIBentryALTinterwordspacing{\spaceskip=\fontdimen2\font plus
\BIBentryALTinterwordstretchfactor\fontdimen3\font minus
  \fontdimen4\font\relax}
\providecommand\BIBforeignlanguage[2]{{%
\expandafter\ifx\csname l@#1\endcsname\relax
\typeout{** WARNING: IEEEtran.bst: No hyphenation pattern has been}%
\typeout{** loaded for the language `#1'. Using the pattern for}%
\typeout{** the default language instead.}%
\else
\language=\csname l@#1\endcsname
\fi
#2}}

\bibitem{ARcore}
Google, ``{ARCore},'' \url{https://developers.google.com/ar}.

\bibitem{ARkit}
Apple, ``{ARKit},'' \url{https://developer.apple.com/augmented-reality/}.

\bibitem{Katragadda2019_AIAA}
S.~Katragadda, B.~A. Mondal, and A.~Deane, ``Stereoscopic mixed reality in
  unmanned aerial vehicle search and rescue,'' in \emph{AIAA Scitech 2019
  Forum}, 2019, p. 0154.

\bibitem{Chen2022IROS}
C.~Chen, Y.~Yang, P.~Geneva, W.~Lee, and G.~Huang, ``Visual-inertial-aided
  online mav system identification,'' in \emph{IEEE/RSJ International
  Conference on Intelligent Robots and Systems (IROS)}, 2022.

\bibitem{Huang2019ICRA}
G.~Huang, ``Visual-inertial navigation: A concise review,'' in \emph{Proc.
  International Conference on Robotics and Automation}, Montreal, Canada, May
  2019.

\bibitem{Lynen2015_RSS}
S.~Lynen, T.~Sattler, M.~Bosse, J.~A. Hesch, M.~Pollefeys, and R.~Siegwart,
  ``Get out of my lab: Large-scale, real-time visual-inertial localization.''
  in \emph{Robotics: Science and Systems}, vol.~1, 2015, p.~1.

\bibitem{Kasyanov2017_IROS}
A.~Kasyanov, F.~Engelmann, J.~St{\"u}ckler, and B.~Leibe, ``Keyframe-based
  visual-inertial online slam with relocalization,'' in \emph{2017 IEEE/RSJ
  international conference on intelligent robots and systems (IROS)}.\hskip 1em
  plus 0.5em minus 0.4em\relax IEEE, 2017, pp. 6662--6669.

\bibitem{Sarlin2019_CVPR}
P.-E. Sarlin, C.~Cadena, R.~Siegwart, and M.~Dymczyk, ``From coarse to fine:
  Robust hierarchical localization at large scale,'' in \emph{CVPR}, 2019.

\bibitem{Geneva2022ICRA}
P.~Geneva and G.~Huang, ``Map-based visual-inertial localization: A numerical
  study,'' in \emph{Proc. International Conference on Robotics and Automation},
  Philadelphia, USA, May 2022.

\bibitem{Mourikis2009_TRO}
A.~I. Mourikis, N.~Trawny, S.~I. Roumeliotis, A.~E. Johnson, A.~Ansar, and
  L.~Matthies, ``Vision-aided inertial navigation for spacecraft entry,
  descent, and landing,'' \emph{IEEE Transactions on Robotics}, 2009.

\bibitem{Schneider2018_RAL}
T.~Schneider, M.~T. Dymczyk, M.~Fehr, K.~Egger, S.~Lynen, I.~Gilitschenski, and
  R.~Siegwart, ``{maplab: An Open Framework for Research in Visual-inertial
  Mapping and Localization},'' \emph{IEEE Robotics and Automation Letters},
  vol.~3, no.~3, pp. 1418--1425, 2018.

\bibitem{Cramariuc2022_RAL}
A.~Cramariuc, L.~Bernreiter, F.~Tschopp, M.~Fehr, V.~Reijgwart, J.~Nieto,
  R.~Siegwart, and C.~Cadena, ``{maplab 2.0 – A Modular and Multi-Modal
  Mapping Framework},'' \emph{IEEE Robotics and Automation Letters}, vol.~8,
  no.~2, pp. 520--527, 2023.

\bibitem{Galvez2012_TRO}
D.~G\'alvez-L\'opez and J.~D. Tard\'os, ``Bags of binary words for fast place
  recognition in image sequences,'' \emph{IEEE Transactions on Robotics}, 2012.

\bibitem{Lynen2014_IC3V}
S.~Lynen, M.~Bosse, P.~Furgale, and R.~Siegwart, ``Placeless
  place-recognition,'' in \emph{2014 2nd International Conference on 3D
  Vision}, vol.~1.\hskip 1em plus 0.5em minus 0.4em\relax IEEE, 2014, pp.
  303--310.

\bibitem{Arandjelovic2016_CVPR}
R.~Arandjelovic, P.~Gronat, A.~Torii, T.~Pajdla, and J.~Sivic, ``{NetVLAD}: Cnn
  architecture for weakly supervised place recognition,'' in \emph{Proceedings
  of the IEEE conference on computer vision and pattern recognition}, 2016, pp.
  5297--5307.

\bibitem{Li2023_JR}
K.~Li, Y.~Ma, X.~Wang, L.~Ji, N.~Geng, \emph{et~al.}, ``Evaluation of global
  descriptor methods for appearance-based visual place recognition,''
  \emph{Journal of Robotics}, vol. 2023, 2023.

\bibitem{Mildenhall2020_ECCV}
B.~Mildenhall, P.~P. Srinivasan, M.~Tancik, J.~T. Barron, R.~Ramamoorthi, and
  R.~Ng, ``Nerf: Representing scenes as neural radiance fields for view
  synthesis.''\hskip 1em plus 0.5em minus 0.4em\relax ECCV, 2020.

\bibitem{Jiang2023_arxiv}
C.~Jiang, H.~Zhang, P.~Liu, Z.~Yu, H.~Cheng, B.~Zhou, and S.~Shen,
  ``H2-mapping: Real-time dense mapping using hierarchical hybrid
  representation,'' 2023.

\bibitem{Chung2023_ICRA}
C.-M. Chung, Y.-C. Tseng, Y.-C. Hsu, X.-Q. Shi, Y.-H. Hua, J.-F. Yeh, W.-C.
  Chen, Y.-T. Chen, and W.~H. Hsu, ``Orbeez-slam: A real-time monocular visual
  slam with orb features and nerf-realized mapping,'' in \emph{2023 IEEE
  International Conference on Robotics and Automation (ICRA)}.\hskip 1em plus
  0.5em minus 0.4em\relax IEEE, 2023, pp. 9400--9406.

\bibitem{Mueller2022_ACM}
T.~M{\"u}ller, A.~Evans, C.~Schied, and A.~Keller, ``Instant neural graphics
  primitives with a multiresolution hash encoding,'' \emph{ACM Transactions on
  Graphics (ToG)}, vol.~41, no.~4, pp. 1--15, 2022.

\bibitem{Chum2003_BMVC}
O.~Chum, J.~Matas, and J.~Kittler, ``Locally optimized ransac,'' in
  \emph{Pattern Recognition: 25th DAGM Symposium, Magdeburg, Germany, September
  10-12, 2003. Proceedings 25}.\hskip 1em plus 0.5em minus 0.4em\relax
  Springer, 2003.

\bibitem{Chum2008_PAML}
O.~Chum and J.~Matas, ``Optimal randomized ransac,'' \emph{IEEE Transactions on
  Pattern Analysis and Machine Intelligence}, 2008.

\bibitem{Sattler2011_ICCV}
T.~Sattler, B.~Leibe, and L.~Kobbelt, ``Fast image-based localization using
  direct 2d-to-3d matching,'' in \emph{2011 International Conference on
  Computer Vision}, 2011.

\bibitem{Liu2017_ICCV}
L.~Liu, H.~Li, and Y.~Dai, ``Efficient global 2d-3d matching for camera
  localization in a large-scale 3d map,'' in \emph{2017 IEEE International
  Conference on Computer Vision (ICCV)}, 2017.

\bibitem{Sattler2017_PAMI}
T.~Sattler, B.~Leibe, and L.~Kobbelt, ``Efficient \& effective prioritized
  matching for large-scale image-based localization,'' \emph{IEEE Transactions
  on Pattern Analysis and Machine Intelligence}, 2017.

\bibitem{Schoenberger2016_CVPR}
J.~L. Schonberger and J.-M. Frahm, ``Structure-from-motion revisited,'' in
  \emph{Proceedings of the IEEE conference on computer vision and pattern
  recognition}, 2016, pp. 4104--4113.

\bibitem{Schoenberger2016_ECCV}
J.~L. Sch\"{o}nberger, E.~Zheng, M.~Pollefeys, and J.-M. Frahm, ``Pixelwise
  view selection for unstructured multi-view stereo,'' in \emph{European
  Conference on Computer Vision (ECCV)}, 2016.

\bibitem{Humenberger2020_ARXIV}
M.~Humenberger, Y.~Cabon, N.~Guerin, J.~Morat, V.~Leroy, J.~Revaud, P.~Rerole,
  N.~Pion, C.~de~Souza, and G.~Csurka, ``Robust image retrieval-based visual
  localization using kapture,'' \emph{arXiv preprint arXiv:2007.13867}, 2020.

\bibitem{Detone2018_CVPR}
D.~DeTone, T.~Malisiewicz, and A.~Rabinovich, ``Superpoint: Self-supervised
  interest point detection and description,'' in \emph{Proceedings of the IEEE
  conference on computer vision and pattern recognition workshops}, 2018.

\bibitem{Kendall2015_ICCV}
A.~Kendall, M.~Grimes, and R.~Cipolla, ``Posenet: A convolutional network for
  real-time 6-dof camera relocalization,'' in \emph{Proceedings of the IEEE
  international conference on computer vision}, 2015, pp. 2938--2946.

\bibitem{Kendall2017_CVPR}
A.~Kendall and R.~Cipolla, ``Geometric loss functions for camera pose
  regression with deep learning,'' in \emph{Proceedings of the IEEE conference
  on computer vision and pattern recognition}, 2017, pp. 5974--5983.

\bibitem{Yang2019_CVPR}
L.~Yang, Z.~Bai, C.~Tang, H.~Li, Y.~Furukawa, and P.~Tan, ``Sanet: Scene
  agnostic network for camera localization,'' in \emph{Proceedings of the
  IEEE/CVF international conference on computer vision}, 2019, pp. 42--51.

\bibitem{Bloesch2015_IROS}
M.~Bloesch, S.~Omari, M.~Hutter, and R.~Siegwart, ``Robust visual inertial
  odometry using a direct ekf-based approach,'' in \emph{2015 IEEE/RSJ
  international conference on intelligent robots and systems (IROS)}.\hskip 1em
  plus 0.5em minus 0.4em\relax IEEE, 2015, pp. 298--304.

\bibitem{Bloesch2017_IJRR}
M.~Bloesch, M.~Burri, S.~Omari, M.~Hutter, and R.~Siegwart, ``Iterated extended
  kalman filter based visual-inertial odometry using direct photometric
  feedback,'' \emph{The International Journal of Robotics Research}, 2017.

\bibitem{Klein2007_ISMAR}
G.~Klein and D.~Murray, ``Parallel tracking and mapping for small {AR}
  workspaces,'' in \emph{Proc. Sixth {IEEE} and {ACM} International Symposiumon
  Mixed and Augmented Reality}, 2007.

\bibitem{Leutenegger2015_IJRR}
S.~Leutenegger, S.~Lynen, M.~Bosse, R.~Siegwart, and P.~Furgale,
  ``Keyframe-based visual--inertial odometry using nonlinear optimization,''
  \emph{The International Journal of Robotics Research}, 2015.

\bibitem{Engel2014_ECCV}
J.~Engel, T.~Sch{\"o}ps, and D.~Cremers, ``Lsd-slam: Large-scale direct
  monocular slam,'' in \emph{European conference on computer vision}.\hskip 1em
  plus 0.5em minus 0.4em\relax Springer, 2014, pp. 834--849.

\bibitem{Forster2014_ICRA}
C.~Forster, M.~Pizzoli, and D.~Scaramuzza, ``{SVO}: Fast semi-direct monocular
  visual odometry,'' in \emph{IEEE International Conference on Robotics and
  Automation (ICRA)}, 2014.

\bibitem{Lin2021_ICCV}
C.-H. Lin, W.-C. Ma, A.~Torralba, and S.~Lucey, ``Barf: Bundle-adjusting neural
  radiance fields,'' in \emph{IEEE International Conference on Computer Vision
  ({ICCV})}, 2021.

\bibitem{Wang2021_arxiv}
Z.~Wang, S.~Wu, W.~Xie, M.~Chen, and V.~Adrian~Prisacariu, ``Nerf$--$: Neural
  radiance fields without known camera parameters,'' \emph{arXiv preprint
  arXiv:2102.07064}, 2021.

\bibitem{Sucar2021_ICCV}
E.~Sucar, S.~Liu, J.~Ortiz, and A.~Davison, ``{iMAP}: Implicit mapping and
  positioning in real-time,'' in \emph{Proceedings of the International
  Conference on Computer Vision ({ICCV})}, 2021.

\bibitem{Zhu2022_CVPR}
Z.~Zhu, S.~Peng, V.~Larsson, W.~Xu, H.~Bao, Z.~Cui, M.~R. Oswald, and
  M.~Pollefeys, ``Nice-slam: Neural implicit scalable encoding for slam,'' in
  \emph{Proceedings of the IEEE/CVF Conference on Computer Vision and Pattern
  Recognition (CVPR)}, 2022.

\bibitem{Rosinol2022_arxiv}
A.~Rosinol, J.~J. Leonard, and L.~Carlone, ``{NeRF-SLAM}: Real-time dense
  monocular slam with neural radiance fields,'' \emph{arXiv preprint
  arXiv:2210.13641}, 2022.

\bibitem{Yen2020_IROS}
L.~Yen-Chen, P.~Florence, J.~T. Barron, A.~Rodriguez, P.~Isola, and T.-Y. Lin,
  ``{iNeRF}: Inverting neural radiance fields for pose estimation,'' in
  \emph{IEEE/RSJ International Conference on Intelligent Robots and Systems
  ({IROS})}, 2021.

\bibitem{Maggio2023_ICRA}
D.~Maggio, M.~Abate, J.~Shi, C.~Mario, and L.~Carlone, ``Loc-nerf: Monte carlo
  localization using neural radiance fields,'' in \emph{IEEE International
  Conference on Robotics and Automation}.\hskip 1em plus 0.5em minus
  0.4em\relax IEEE, 2023.

\bibitem{Adamkiewicz2022_RAL}
M.~Adamkiewicz, T.~Chen, A.~Caccavale, R.~Gardner, P.~Culbertson, J.~Bohg, and
  M.~Schwager, ``Vision-only robot navigation in a neural radiance world,''
  \emph{IEEE Robotics and Automation Letters}, 2022.

\bibitem{Qin2018_TRO}
T.~Qin, P.~Li, and S.~Shen, ``Vins-mono: A robust and versatile monocular
  visual-inertial state estimator,'' \emph{IEEE Transactions on Robotics},
  2018.

\bibitem{Mourikis2007_ICRA}
A.~I. Mourikis and S.~I. Roumeliotis, ``A multi-state constraint kalman filter
  for vision-aided inertial navigation,'' in \emph{Proceedings 2007 IEEE
  International Conference on Robotics and Automation}, 2007.

\bibitem{Geneva2020ICRA}
\BIBentryALTinterwordspacing
P.~Geneva, K.~Eckenhoff, W.~Lee, Y.~Yang, and G.~Huang, ``{OpenVINS:} a
  research platform for visual-inertial estimation,'' in \emph{Proc. of the
  IEEE International Conference on Robotics and Automation}, Paris, France,
  2020. [Online]. Available: \url{https://github.com/rpng/open_vins}
\BIBentrySTDinterwordspacing

\bibitem{Chatfield1997}
A.~B. Chatfield, \emph{Fundamentals of High Accuracy Inertial
  Navigation}.\hskip 1em plus 0.5em minus 0.4em\relax Reston, VA: American
  Institute of Aeronautics and Astronautics, Inc., 1997.

\bibitem{Geneva2019ICRA}
P.~Geneva, K.~Eckenhoff, and G.~Huang, ``A linear-complexity {EKF} for
  visual-inertial navigation with loop closures,'' in \emph{Proc. International
  Conference on Robotics and Automation}, Montreal, Canada, May 2019.

\bibitem{Dong2016_ECCV}
C.~Dong, C.~C. Loy, and X.~Tang, ``Accelerating the super-resolution
  convolutional neural network,'' in \emph{Computer Vision--ECCV 2016: 14th
  European Conference, Amsterdam, The Netherlands, October 11-14, 2016,
  Proceedings, Part II 14}.\hskip 1em plus 0.5em minus 0.4em\relax Springer,
  2016.

\bibitem{TensorRT}
NVIDIA, ``{TensorRT},'' \url{https://github.com/NVIDIA/TensorRT}.

\bibitem{Bradski2000}
G.~Bradski, ``{The OpenCV Library},'' \emph{Dr. Dobb's Journal of Software
  Tools}, 2000.

\bibitem{alcantarilla2011fast}
P.~F. Alcantarilla and T.~Solutions, ``Fast explicit diffusion for accelerated
  features in nonlinear scale spaces,'' \emph{IEEE Trans. Patt. Anal. Mach.
  Intell}, vol.~34, no.~7, pp. 1281--1298, 2011.

\bibitem{alcantarilla2012kaze}
P.~F. Alcantarilla, A.~Bartoli, and A.~J. Davison, ``{KAZE} features,'' in
  \emph{Computer Vision--ECCV 2012: 12th European Conference on Computer
  Vision, Florence, Italy, October 7-13, 2012, Proceedings, Part VI 12}.\hskip
  1em plus 0.5em minus 0.4em\relax Springer, 2012, pp. 214--227.

\bibitem{leutenegger2011brisk}
S.~Leutenegger, M.~Chli, and R.~Y. Siegwart, ``{BRISK}: Binary robust invariant
  scalable keypoints,'' in \emph{2011 International conference on computer
  vision}.\hskip 1em plus 0.5em minus 0.4em\relax IEEE, 2011, pp. 2548--2555.

\bibitem{rublee2011orb}
E.~Rublee, V.~Rabaud, K.~Konolige, and G.~Bradski, ``{ORB}: An efficient
  alternative to sift or surf,'' in \emph{2011 International conference on
  computer vision}.\hskip 1em plus 0.5em minus 0.4em\relax Ieee, 2011, pp.
  2564--2571.

\bibitem{Zhang2018_IROS}
Z.~Zhang and D.~Scaramuzza, ``A tutorial on quantitative trajectory evaluation
  for visual (-inertial) odometry,'' in \emph{2018 IEEE/RSJ International
  Conference on Intelligent Robots and Systems (IROS)}.\hskip 1em plus 0.5em
  minus 0.4em\relax IEEE, 2018, pp. 7244--7251.

\bibitem{Chen2023ICRA}
\BIBentryALTinterwordspacing
C.~Chen, P.~Geneva, Y.~Peng, W.~Lee, and G.~Huang, ``Monocular visual-inertial
  odometry with planar regularities,'' in \emph{Proc. of the IEEE International
  Conference on Robotics and Automation}, London, UK., 2023. [Online].
  Available: \url{https://github.com/rpng/ar_table_dataset}
\BIBentrySTDinterwordspacing

\bibitem{He2016_CVPR}
K.~He, X.~Zhang, S.~Ren, and J.~Sun, ``Deep residual learning for image
  recognition,'' in \emph{Proceedings of the IEEE conference on computer vision
  and pattern recognition}, 2016.

\bibitem{Katragadda2023TRNERF}
S.~Katragadda, W.~Lee, Y.~Peng, P.~Geneva, C.~Chen, and G.~Huang,
  ``{NeRF-VINS}: A real-time neural radiance field map-based visual-inertial
  navigation system,'' University of Delaware, Tech. Rep. RPNG-2023-NERF, 2023,
  available: \url{http://udel.edu/~ghuang/papers/tr_nerf.pdf}.

\bibitem{Geneva2020TRVICON2GT}
P.~Geneva and G.~Huang, ``vicon2gt: Derivations and analysis,'' University of
  Delaware, Tech. Rep. RPNG-2020-VICON2GT, 2020, available:
  \url{http://udel.edu/~ghuang/papers/tr_vicon2gt.pdf}.

\bibitem{Lindenberger2023_ICCV}
P.~Lindenberger, P.-E. Sarlin, and M.~Pollefeys, ``{LightGlue: Local Feature
  Matching at Light Speed},'' in \emph{ICCV}, 2023.

\bibitem{qin2019general}
T.~Qin, S.~Cao, J.~Pan, and S.~Shen, ``A general optimization-based framework
  for global pose estimation with multiple sensors,'' \emph{arXiv preprint
  arXiv:1901.03642}, 2019.

\bibitem{Wang2023_CVPR}
P.~Wang, Y.~Liu, Z.~Chen, L.~Liu, Z.~Liu, T.~Komura, C.~Theobalt, and W.~Wang,
  ``{F2-NeRF}: Fast neural radiance field training with free camera
  trajectories,'' in \emph{Proceedings of the IEEE/CVF Conference on Computer
  Vision and Pattern Recognition}, 2023, pp. 4150--4159.

\bibitem{Tancik2022_CVPR}
M.~Tancik, V.~Casser, X.~Yan, S.~Pradhan, B.~Mildenhall, P.~P. Srinivasan,
  J.~T. Barron, and H.~Kretzschmar, ``{Block-NeRF}: Scalable large scene neural
  view synthesis,'' in \emph{Proceedings of the IEEE/CVF Conference on Computer
  Vision and Pattern Recognition}, 2022, pp. 8248--8258.

\bibitem{Kerbl2023_ACM}
B.~Kerbl, G.~Kopanas, T.~Leimk{\"u}hler, and G.~Drettakis, ``3d gaussian
  splatting for real-time radiance field rendering,'' \emph{ACM Transactions on
  Graphics}, vol.~42, no.~4, 2023.

\end{thebibliography}
}

\end{document}